\documentclass{article}
\usepackage{ijcai20}

\usepackage{times}
\usepackage{soul}
\usepackage{url}
\usepackage[hidelinks]{hyperref}
\usepackage[utf8]{inputenc}
\usepackage[small]{caption}
\usepackage{graphicx}
\usepackage{amsmath}
\usepackage{amsthm}
\usepackage{booktabs}
\usepackage{algorithm}
\usepackage{algorithmic}
\usepackage{subfigure}
\usepackage{color}
\title{HRDNet: High-resolution Detection Network for Small Objects}

\author{
Ziming Liu$^1$
\and
Guangyu Gao$^1$\and
Lin Sun$^{2}$\And
Zhiyuan Fang$^1$
\affiliations
$^1$Beijing Institute of Technology\\
$^2$SAMSUNG SEMICONDUCTOR INC. \\
}

\begin{document}
\maketitle

\begin{abstract}
Small object detection is challenging because small objects do not contain detailed information and may even disappear in the deep network. Usually, feeding high-resolution images into a network can alleviate this issue. However, simply enlarging the resolution will cause more problems, such as that, it aggravates the large variant of object scale and introduces unbearable computation cost. To keep the benefits of high-resolution images without bringing up new problems, we proposed the High-Resolution Detection Network (HRDNet). HRDNet takes multiple resolution inputs using multi-depth backbones. To fully take advantage of multiple features, we proposed Multi-Depth Image Pyramid Network (MD-IPN) and Multi-Scale Feature Pyramid Network (MS-FPN) in HRDNet. MD-IPN maintains multiple position information using multiple depth backbones. Specifically, high-resolution input will be fed into a shallow network to reserve more positional information and reducing the computational cost while low-resolution input will be fed into a deep network to extract more semantics. By extracting various features from high to low resolutions, the MD-IPN is able to improve the performance of small object detection as well as maintaining the performance of middle and large objects. MS-FPN is proposed to align and fuse multi-scale feature groups generated by MD-IPN to reduce the information imbalance between these multi-scale multi-level features. Extensive experiments and ablation studies are conducted on the standard benchmark dataset MS COCO2017, Pascal VOC2007/2012 and a typical small object dataset, VisDrone 2019. Notably, our proposed HRDNet achieves the state-of-the-art on these datasets and it performs better on small objects. 
\end{abstract}

\section{Introduction}
\label{intr}
Object detection is challenging while it has widespread applications. With the advances of deep learning, object detection achieves the remarkable progress.
According to whether the proposals are generated by an independent learning stage or directly and densely sample possible locations, object detection can be classified into two-stage or one-stage models. Compared to two-stage detectors \cite{cai2018cascade,fasterrcnn} one stage methods \cite{retina,SSD} are less complex, therefore, it can run faster with some precision loss. While most existed successful methods are based on anchor mechanism, the recent state-of-the-art methods focus on anchor-free detection mostly, e.g. CornerNet  \cite{CornerNet}, FCOS  \cite{tian2019fcos}, FSAF \cite{fsaf}. These CNN based detection methods are very powerful because it can create some low-level abstractions of the images like lines, circles and then ‘iteratively combine’ them into some objects, but this is also the reason that they struggle with detecting small objects. 

\begin{figure*}[t]
    \centering
    \includegraphics[width=1.0\textwidth, height=6.5cm]{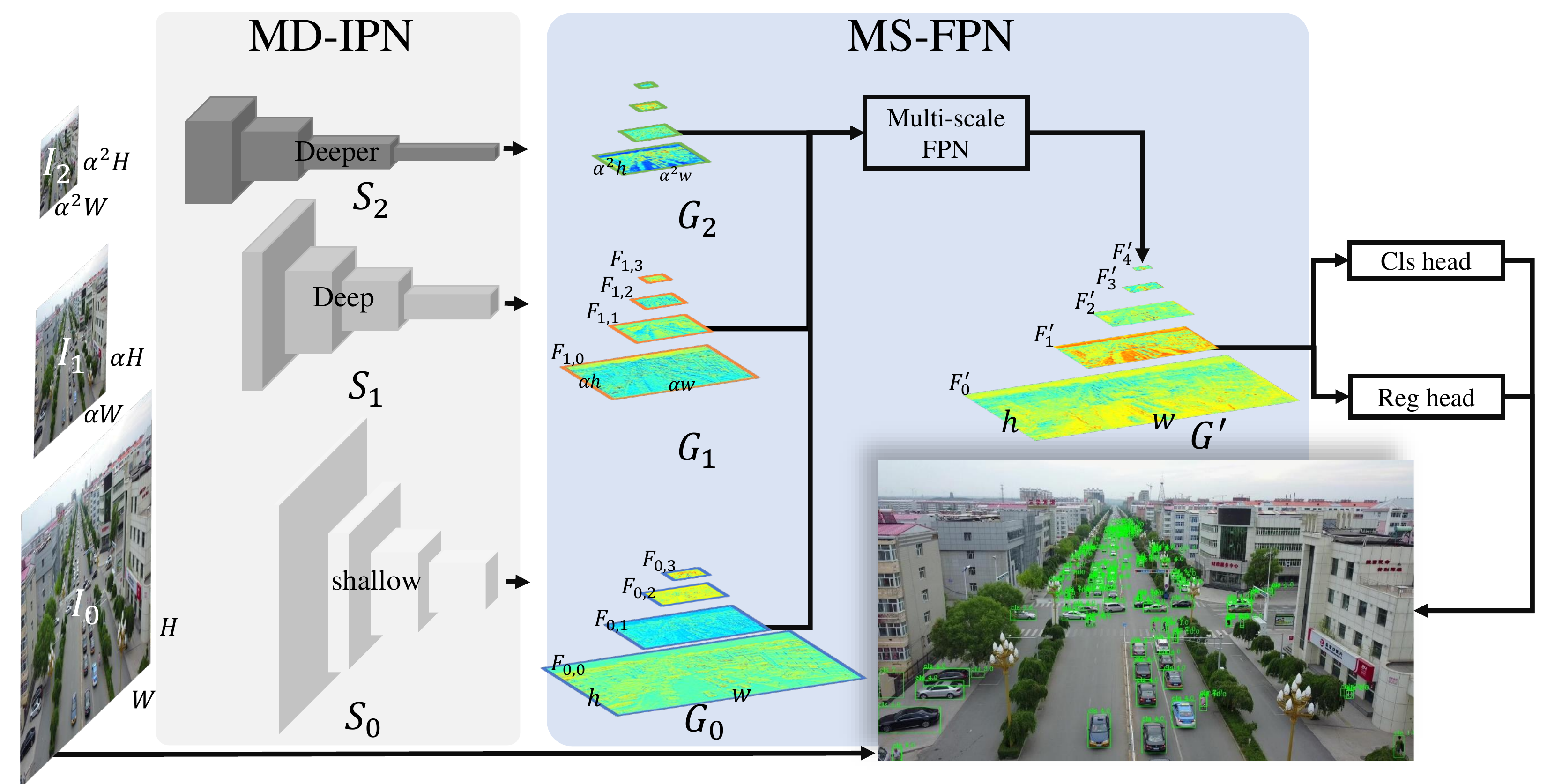}
    \caption{The overall structure of HRDNet, composed by MD-IPN and MS-FPN. The input of HRDNet is an image pyramid with $N$ images, $N=3$ in this figure, the decreasing ratio is $\alpha$. The outputs of MD-IPN are $N$ groups feature pyramids, the decreasing ratio of each feature pyramid is $2$. MS-FPN fuses these features into single one feature pyramid $\{F^{'}_{0},F^{'}_{1},F^{'}_{2},F^{'}_{3},F^{'}_{4}\}$, which is used for object detection. The exemplar comes from VisDrone2019 validation set.}
    \label{fig:mainfig}
\end{figure*}

Generally, the object detection algorithms mentioned above can achieve good performance, as long as the features extracted from the backbone network are strong enough. Usually, a huge and deep CNN backbone extracts multi-level features and then refine them with feature pyramid network (FPN). Most time, these detection models benefit from deeper backbone, while the deeper backbone also introduces more computation cost and memory usage.

Commonly, the detection performance is extremely sensitive to the resolution of input. High-resolution images are more suitable for small object detection, which reserves more details and position information. But high-resolution also introduces new problems, such as, (i) it's easy to damage the detection of large objects, as shown in Table \ref{tab:prove_size}; (ii) Detection always needs a deeper network for more powerful semantics, resulting in an unaffordable computing cost. Actually, it's essential to use the high-resolution image for small object detection, and also the deeper backbone for small scale images. But we should deal with the trade-offs between large and small object detection, as well as high performance and low computational complexity.

To solve these problems, we propose a new architecture, High-resolution Detection Network (HRDNet). As shown in Figure \ref{fig:mainfig}, it includes two parts: Multi-Depth Image Pyramid Network (MD-IPN) and Multi-Scale Feature Pyramid Network (MS-FPN). The main idea of the HRDNet is to use a deep backbone to process low-resolution images while using a shallow backbone to process high-resolution images. The advantage of extracting features from high-resolution images with the shallow and tiny network has been demonstrated in \cite{doubleRtinyobj}. With HRDNet, we can not only get more details for a small object in high-resolution, but also guarantee the efficiency and effectiveness by integrating multi-depth and multi-scale deep networks.

MD-IPN can be regarded as a variant of the image pyramid network with multiple streams, as shown in Figure \ref{fig:mainfig}. MD-IPN is dealing with the trade-offs between large and small object detection, as well as high performance and low computational complexity. We extract features from the high-resolution image using a shallow backbone network. Because of the weak semantic representation power of the shallow backbone network, we also need deep backbones to obtain semantically strong features by feeding low-resolution images in. Thus, the inputs of the MD-IPN form an image pyramid with a fixed decreasing ratio of $\alpha \in [0,1]$. The output of MD-IPN is a series of multi-scale feature groups, and each group contains multi-level feature maps.

The multi-scale feature groups extend the standard feature pyramid by adding multi-scale streams. Therefore, traditional FPN can't be directly applied here. To fuse these multi-scale feature groups properly, we proposed the Multi-Scale Feature Pyramid Network (MS-FPN). As shown in Figure \ref{fig:multiscalefpn}, the information of images not only propagates from high-level features to low-level features inside the multi-level feature pyramid but also between streams of different scales in MD-IPN.


Before going through the details, we summarize our contributions as follows:
\begin{itemize}
\item  We comprehensively analyzed the factors that small object detection depends on and the trade-off between performance and efficiency, as well as proposed a novel high-resolution detection network, HRDNet, considering both image pyramid and feature pyramid.
\item  In HRDNet, we designed the multi-depth and multi-stream module, MD-IPN to balance the performance between small, middle and large objects. We proposed another new module, MS-FPN to combine different semantic representations from these multi-scale feature groups.
\item Extensive ablation studies validate the effectiveness and efficiency of the proposed approach. The performance of bench-marking on several datasets show that our approach achieves the state-of-the-art performance on object detection, particularly on small object detection. Meanwhile, we hope such practice of small object detection could shed the light for other researches.
\end{itemize}
\vspace{-0.2cm}
\section{Related Work}
Object detection is a basic task for many downstream tasks in computer vision. The state-of-the-art detection networks include one stage model, e.g., RetinaNet \cite{retina}, Yolo-v3 \cite{redmon2018yolov3}, Center net  \cite{duan2019centernet}, FSAF \cite{fsaf}, Corner net \cite{CornerNet} and two-stage model, e.g., Faster R-CNN  \cite{fasterrcnn}, Cascade R-CNN \cite{cai2018cascade} etc.).
Nevertheless, the proposed HRDNet is a more fundamental framework that could be the backbone network for most of the detection models, as mentioned above, such as RetinaNet and Cascade R-CNN.

\paragraph{Small object detection}
The detection performance is largely restricted by small object detection in most datasets. Therefore, there are many researches specializing in small object detection. For example, \cite{Augmentation} proposed oversampling and copy-pasting small objects to solve such a problem.
Perceptual GAN \cite{Perceptual_GAN} generated super-resolved features and stacked them into feature maps of small objects to enhance the representations. DetNet \cite{detnet} maintained the spatial resolution and has a large receptive field to improve small object detection. SNIP \cite{SNIP} resized images to different resolutions and only train samples which is close to ground truth. SNIPER \cite{SNIPER} is proposed to use regions around the bounding box to remove the influence of background. Unlike these methods, we combine both image pyramid and feature pyramid together, with which it not only effectively improves the detection performance of small targets, but also ensure the detection performance of other objects.


\paragraph{High-resolution detection}
Some studies already explored to do object detection on high-resolution images. \cite{doubleRtinyobj} proposed a fast tiny detection network for high-resolution remote sensing images. \cite{4k8kdetection} proposed an attention pipeline to achieve fast detection on 4K or 8K videos using YOLO v2 \cite{yolov2}. However, these works did not fully explore the effect of high-resolution images for small object detection, which is what we concentrate on.

\paragraph{Feature-level imbalance}
To capture the semantic information of objects from different scales, multi-level features are commonly used for object detection. However, they have serious feature-level imbalance because they convey different semantic information. Feature Pyramid Network (FPN) \cite{FPN} introduced a top-down pathway to transmit semantic information, alleviating the feature imbalance problem in some degree. Based on FPN, PANet \cite{PANet} involved a bottom-up path to enrich the location information of deep layers. The authors of Libra R-CNN \cite{Libra_RCNN} revealed and tried to deal with the sample level, feature level, and objective level imbalance issues. Pang et al. \cite{Pang_2019_CVPR} proposed a light weighted module to produce featured image pyramid features to augment the output feature pyramid. While these methods only focus on multi-level features. Here, We solve the feature-level imbalance from a new angle, we proposed a new module called Multi-scale FPN to solve the imbalance not only from multi-level features but also from multi-scale feature groups.

\vspace{-0.2cm}
\section{High-Resolution Detection Network}

Obviously, high-resolution images are important for small object detection. Unfortunately, high-resolution images will introduce unaffordable computation costs to deep networks. At the same time, high-resolution images aggravate the variance of object scales, worsening the performance of large objects, as shown in Table \ref{tab:prove_size}. To balance computation costs and variance of objects scales while keeping the performance of all the classes, we proposed the High-Resolution Detection Network (HRDNet). The HRDNet is a general concept that is compatible with any alternative detection method.

\begin{figure}[h]
    \centering
    \includegraphics[width=0.5\textwidth]{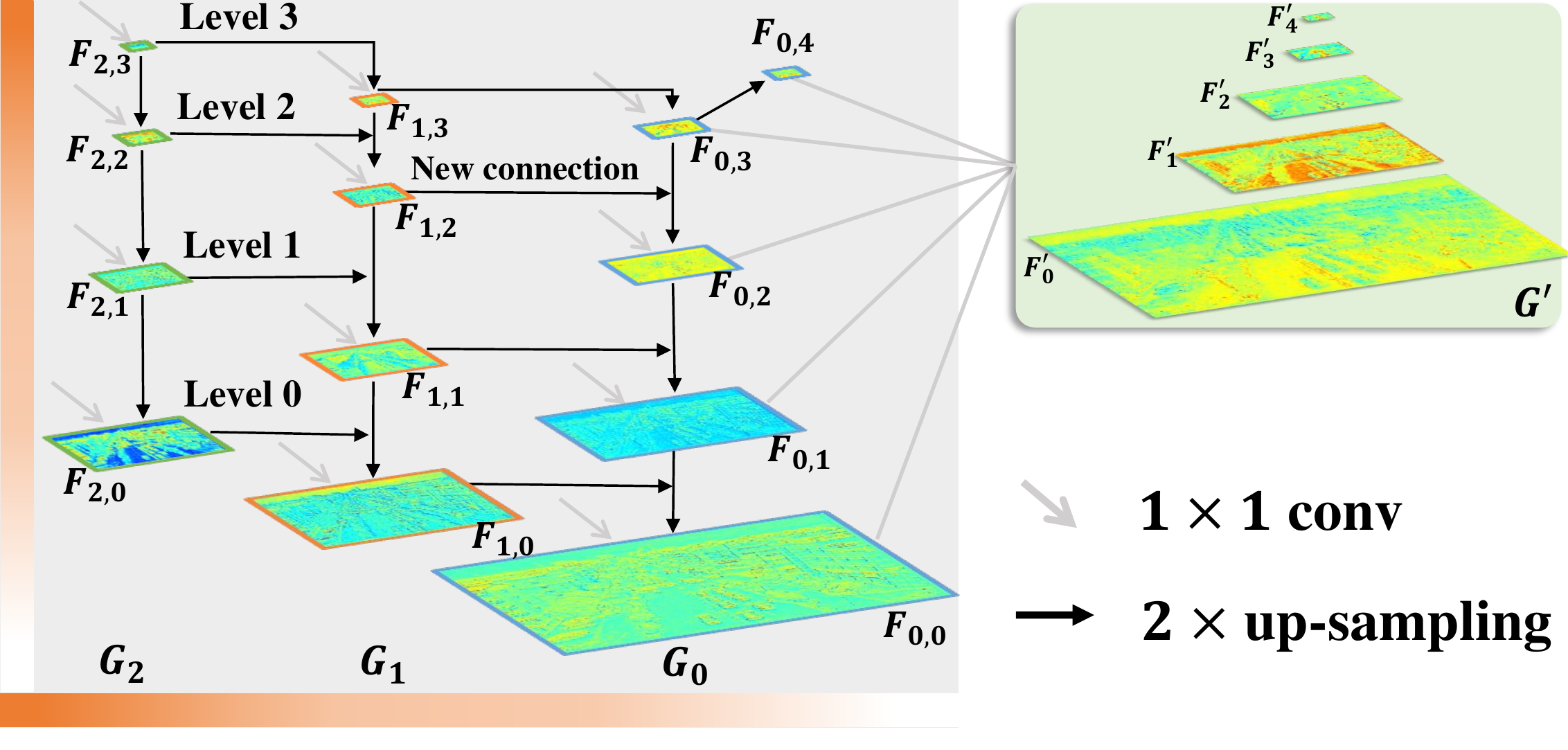}
    \caption{The details of MS-FPN, in which there is a feature pyramid with three streams and four levels. The horizontal orange bar indicates the depth of $\mathcal{S}_{i}$, the vertical orange bar indicates the depth of a single backbone. Better to be viewed in color and zoom in. }
    \label{fig:multiscalefpn}
\end{figure}

More specifically, HRDNet is designed with two novel modules, Multi-Depth Image Pyramid Network (MD-IPN) and Multi-Scale Feature Pyramid Network (MS-FPN). In MD-IPN, an image pyramid is processed by backbones with different depth, i.e., using deep CNNs for the low-resolution images while using shallow CNNs for the high-resolution images, as shown in Figure \ref{fig:mainfig}. After that, to fuse the multi-scale groups of multi-level features from MD-IPN, MS-FPN is proposed as a more reasonable feature pyramid architecture (Figure \ref{fig:multiscalefpn}).

\vspace{-0.2cm}
\subsection{MD-IPN} 

The MD-IPN is composed of $N$ independent backbones with various depth to process the image pyramid. We term each backbone as a \textit{stream}. HRDNet can be generalized to more streams, but to better illustrate the main idea, we mainly discuss the two-stream HRDNet and three-stream HRDNet. Figure \ref{fig:mainfig} presents an example of three-stream HRDNet. Given an image $I$ with resolution $R$, the high-resolution image ($I_0$ with $R$) is processed by a stream of shallow CNN ($S_0$), the lower-resolution images ($I_1$ and $I_2$ with $\alpha R$ and $\alpha^2R$, and $\alpha=0.5$.) is processed by streams of deeper CNN ($S_1$ and $S_2$). Generally, we can build an image pyramid network with $N$ independent parallel streams, $\mathcal{S}_{i}, i=\{0,1,2,...,N-1\}$.

We use $\{I_{i}\}_{i=0}^{N-1}$ to represent the input images with different resolutions given the \textit{original image} $I_{0}$ with the highest resolution. The outputs of the multi scale image pyramid are $N$ feature groups $\{\mathcal{G}_{i}\}_{i=0}^{N-1}$. Each group $\mathcal{G}_i$ contains a set of multi-level features $\{F_{i,j}\}$, where $i \in \{0,1,2,...,N-1\}$ is the multi-scale index and $j \in \{0,1,2,...,M-1\}$ is the multi-level index. For example, in Figure \ref{fig:mainfig}, the value of $N$ and $M$ are $3, 4$, respectively, and the relation can be formulated as 
\begin{equation}
\label{ipnet}
    \mathcal{G}_{i} = \mathcal{S}_{i}(I_{i}) = \{F_{i,0}, F_{i,1}, F_{i,2}, F_{i,3}\},
\end{equation}  
where $i\in\left\{0,1,2,\cdots,N-1\right\}$. 
\vspace{-0.3cm}
\vspace{-0.3cm}
\vspace{-0.2cm}
\subsection{MS-FPN}
\label{ms-fpn}
Feature pyramid network (FPN) is one of the key components for most object detection algorithms. It combines low-resolution, semantically strong features with high-resolution, semantically weak features via a top-down pathway and lateral connections. 

\begin{table*}[ht]
    \centering
    \tiny
    \resizebox{\textwidth}{11mm}{
    \begin{tabular}{c|c|c|c|c|c|c|c|c|c|c|c|c}
    \toprule
    model&resolution&pedestrian&people&bicycle&car&van&truck&tricycle&awning-tri&bus&motor&\textbf{mAP}\\
    \midrule
    Cascade R-CNN&$1333\times800$&37.9&27.7&13.3&74.3&44.6&{\color{blue}{34.7}}&24.6&{\color{blue}{13.2}}&{\color{blue}{52.4}}&38.3&\textbf{36.1}\\
    Cascade R-CNN&$2666\times1600$&51.5&38.0&20.2&80.0&48.0&{\color{blue}{32.4}}&28.2&{\color{blue}{12.1}}&{\color{blue}{44.8}}&47.5&\textbf{40.3}\\
    \hline
    HRDNet&$2000\times1200$&49.6&37.2&17.4&79.8&47.9&{\color{red}{36.9}}&30.4&{\color{red}{15.3}}&{\color{red}{56.0}}&48.7&\textbf{41.9}\\
    HRDNet&$2666\times1600$&55.8&42.4&23.1&82.4&51.2&{\color{red}{42.1}}&34.3&{\color{red}{16.3}}&{\color{red}{59.7}}&53.8&{\color{magenta}{\textbf{46.1}}}\\
    HRDNet $\dagger$&$2666\times1600$&56.7&45.1&27.7&82.6&51.3&43.0&37.6&18.8&58.9&56.4&{\color{red}{\textbf{47.8}}}\\
    \bottomrule
    \end{tabular}}
    \caption{\textbf{Performance Comparison of Cascade R-CNN and HRDNet with different resolution's input.} The HRDNet here is a two streams version, and $\dagger$ means that it is trained on patch images as mentioned in Subsection \ref{technical_details} experiment details.}
    \label{tab:prove_size}
\end{table*}
\begin{figure*}[t]
    \centering
    \subfigure[]{
        \centering
        \includegraphics[width=0.32\linewidth]{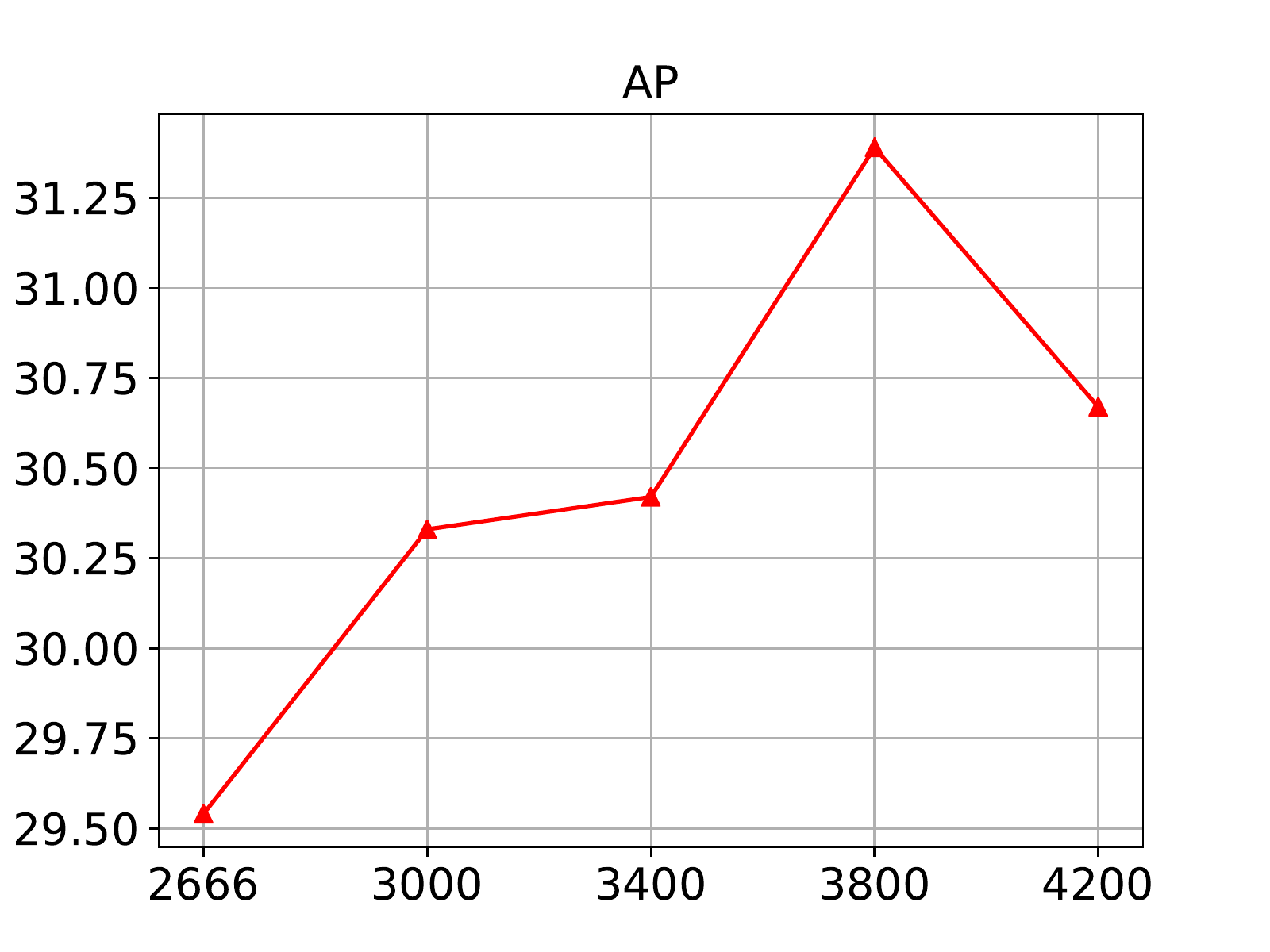}
    }
    \subfigure[]{
        \centering
        \includegraphics[width=0.32\linewidth]{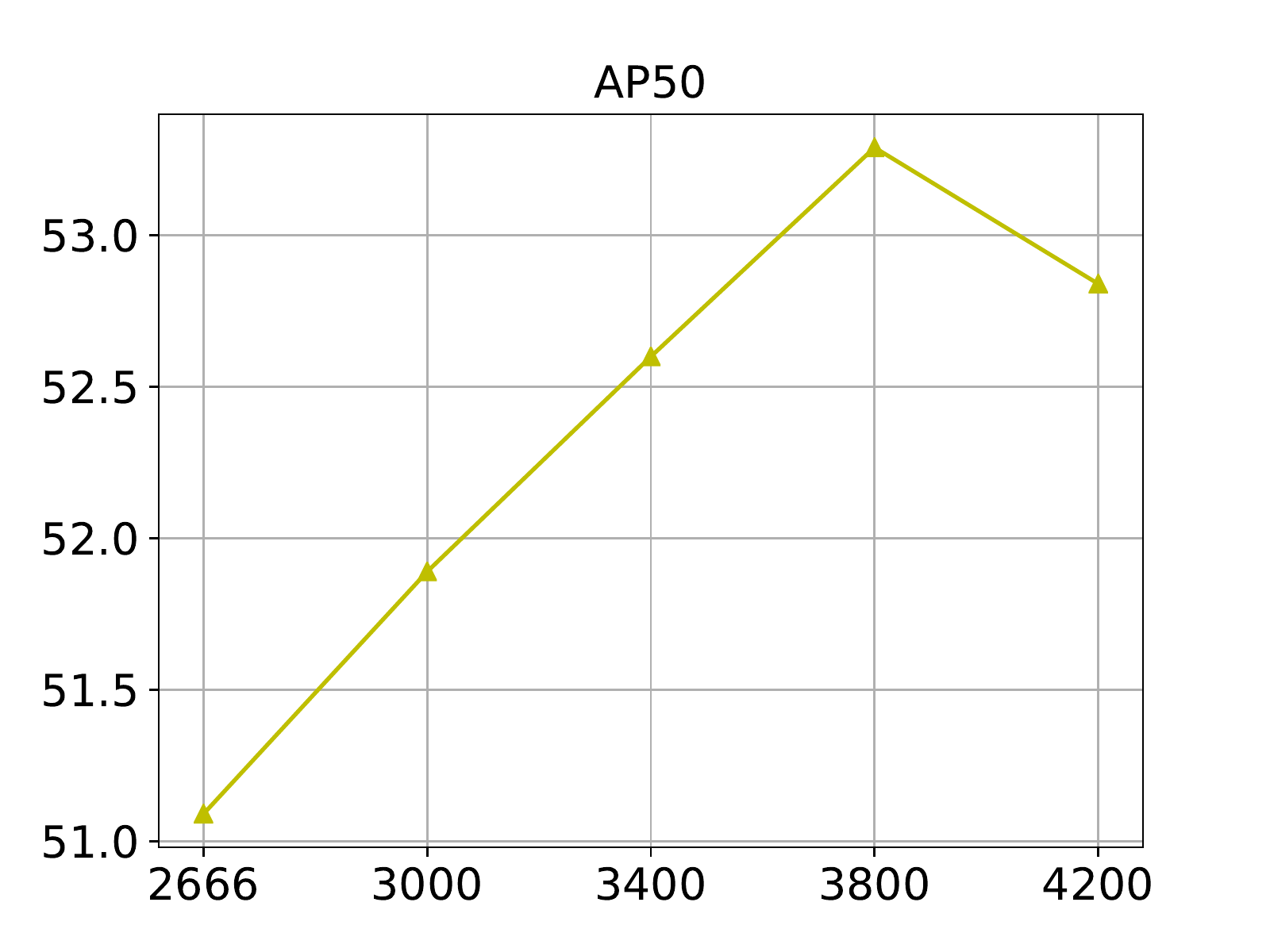}
    }
    \subfigure[]{
        \centering
        \includegraphics[width=0.32\linewidth]{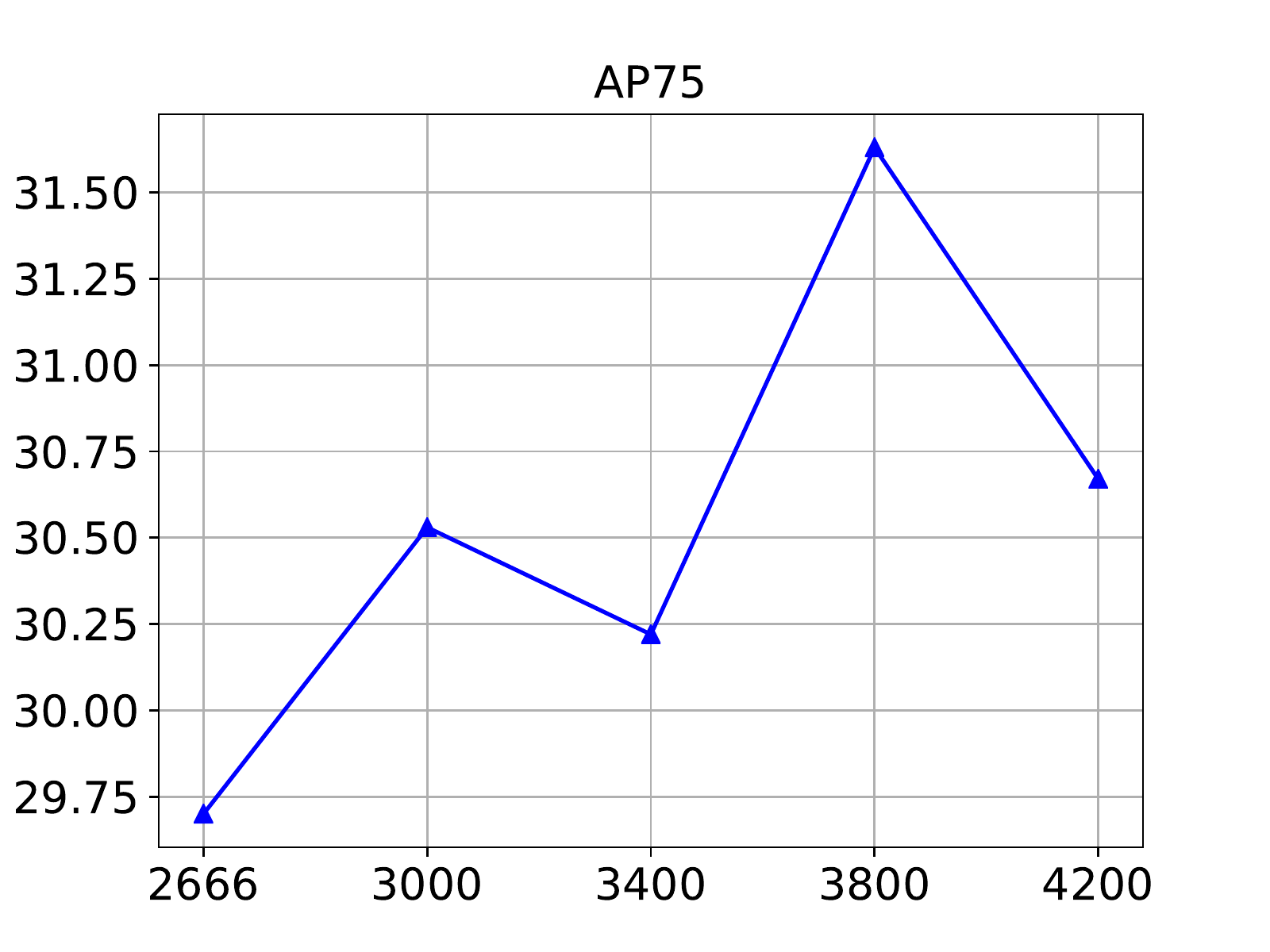}
    }

    \caption{\textbf{The change of $AP, AP50,  AP75$ over different input's resolution.} The HRDNet used here is a two-stream version with ResNet18+101 backbone. The training details follow the subsection \ref{technical_details}.}
    \label{fig:optim_size}
\end{figure*}
In our proposed HRDNet, the MD-IPN generates multi-scale (different resolution) and multi-level (different hierarchy of features) features. To deal with the multi-scale hierarchy features, we also proposed the Multi-Scale FPN (MS-FPN). Different from FPN, semantic information propagates not only from high-level features to low-level features but also from deep stream (low-resolution) to shallow stream (high-resolution). Therefore, there are two directions for the computation of the multi-scale FPN. The basic operation in multi-scale FPN is same as traditional FPN, i.e., $1\times1$Convolution, $2\times$ up-sampling and sum-up.

In this way, the highest resolution feature, i.e., $F_{0,0}$, not only maintain the high-resolution for small object detection but also combine semantically strong features from multi-scale streams. Our novel MS-FPN can be formulated as 

\begin{equation}
    F_{i,j} =
    \begin{cases}
        Conv(F_{i,j}) +  Up(F_{i,j-1})\! & i =N$-$1\\\\
        Conv(F_{i,j}) +  Up(F_{i,j-1}) +  Up(F_{i+1,j})\! & i \neq N$-$1
    \end{cases}
\end{equation}

The $F_{i,j}$ is the feature in level $j$ and stream $i$ in Figure \ref{fig:multiscalefpn}. The $Up(.)$ operation is $2\times$ up-sampling. The $Conv(.)$ is $1\times 1$ convolution.  
Finally, MS-FPN outputs the final feature group $G^{'}=\{F^{'}_{0},F^{'}_{1},...F^{'}_{i},...\}$. $F^{'}_{i}$ is calculated by
\begin{equation}
    F^{'}_{i} = Conv(F_{0,i})
\end{equation} 
where $F_{0,i}$ is the features in Group $G_{0}$, i.e., the outputs of the highest resolution stream.

\vspace{-0.2cm}
\section{Experiments}
\subsection{Experiment details}
\label{technical_details}
\vspace{-0.1cm}
\paragraph{Datasets}
We conduct experiments on both the typical small object detection data set, VisDrone2019 \cite{visdronevid} and traditional datasets of MS COCO2017 \cite{coco} and Pascal VOC2007/2012 \cite{pascalvoc} as well. 

The VisDrone2019 dataset is collected by the AISKYEYE team, which consists of 288 video clips formed by 261,908 frames and 10,209 static images, covering a wide range location, environment, objects, and density. The resolution of VisDrone2019 is higher than COCO as we mentioned in Section \ref{intr}, ranging from 960 to 1360. MS COCO and Pascal VOC are the most common benchmark for object detection. Following common practice, we trained our model on the COCO training set and tested it on the COCO validation set. For Pascal VOC, we trained our model with all the training and validation datasets from both Pascal VOC 2007 and 2012, tested the model on the Pascal VOC 2007 test set. 

\begin{figure}[]
    \centering
    \includegraphics[width=0.4\textwidth]{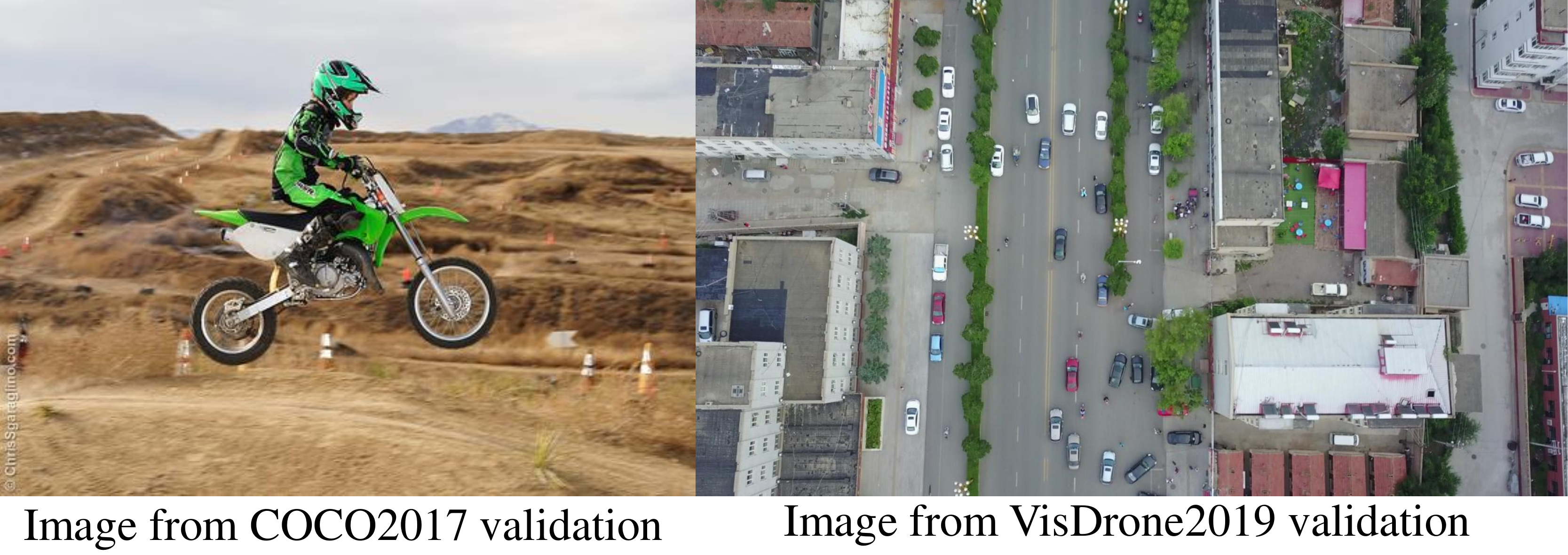}
    \caption{The exemplar from COCO2017 and VisDrone2019.}
    \label{fig:comparedata}
\end{figure}

In COCO or Pascal VOC dataset, most images' resolution is $500$-$800$ px, which is resized to $1333\times800$ or $1000\times600$ in the training stage, but $960$-$1360$ px in VisDrone2019 \cite{visdronedet} dataset. As shown in Figure \ref{fig:comparedata}, compared to MS-COCO, there are more objects and nearly all of them are very small in VisDrone2019, which is more challenging. 
\vspace{-0.1cm}
\paragraph{Training}
We followed the common practice in mmdetection \cite{chen2019mmdetection}. We trained the models on VisDrone2019 with four Nvidia $2080Ti$ GPUs and COCO with eight Nvidia $P100$ GPUs. We use SGD optimizer with a mini-batch $2$ for each GPU. The learning rate starts from 0.02 and decreases by $10$ at epoch $7$ and $11$. The weight decay is $1\times10^{-4}$. The linear warm-up strategy is used with warm-up iterations of $500$, and the warm-up ratio of $1.0/3$. The image pyramid is obtained by the linear interpolation. The resolution decreasing ratio $\alpha$ is $0.5$. 

In order to fit the high-resolution images from VisDrone2019 into GPU memory, we equally cropped each original image in VisDrone2019 training set into four patch images which are not overlapped. In this way, we obtained a new training set with such cropped images.
\vspace{-0.1cm}
\paragraph{Inference}
Same resolution as training is used for inference. The IOU threshold of NMS is $0.5$, and the threshold of confidence score is $0.05$. Without especially emphasizing, for the multi-scale test in our experiments, we use three scales. 
\vspace{-0.2cm}
\subsection{Ablation Studies}
\begin{figure}[]
    \centering
    \includegraphics[width=0.45\textwidth]{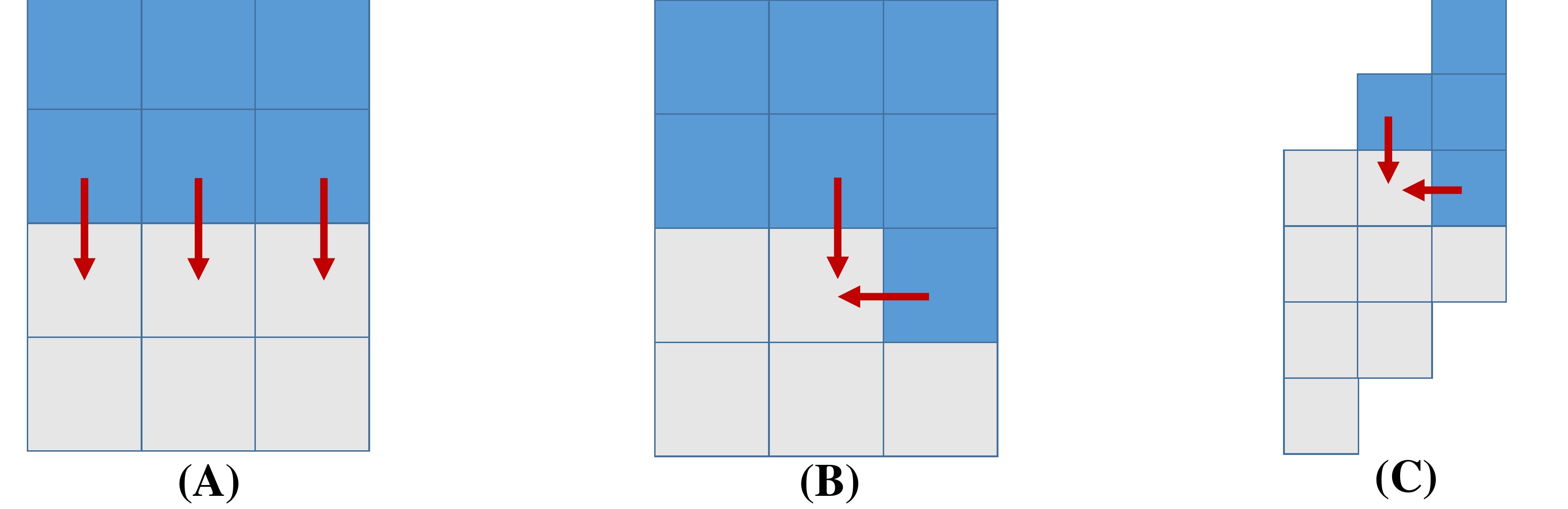}
    \caption{Comparison of the simple FPN, multi-scale FPN aligned with depth and resolution. Each column is one stream in MD-IPN, each row is corresponding to the depth of backbone. The blue blocks represent those features have been fused while gray blocks are those waiting to be fused. The {\color{black}{red arrows}} represent a basic fusing operation described in subsection \ref{ms-fpn}.}
    \label{fig:compareFPN}
\end{figure}

\begin{table}[h]
    \centering
    \tiny
    \setlength{\tabcolsep}{6.5mm}
    \begin{tabular}[]{c|c|c|c}
         \toprule
         style&AP&AP50&AP75\\
         \midrule
         ResNet10+18\\
         \hline
         simple FPN&28.8&49.5&28.8 \\
         aligned by resolution&28.7&49.6&28.7\\
         aligned by depth&\textbf{28.9}&\textbf{49.9}&28.7\\
         \hline
         ResNet18+101\\
         \hline
         
         aligned by resolution&31.8&54.0&32.3\\
         aligned by depth&\textbf{32.0}&\textbf{54.3}&\textbf{32.5}\\ 
         \bottomrule
    \end{tabular}
    \caption{\textbf{The comparison of three different styles of MS-FPN.} }
    \label{tab:multiscaleFPN}
\end{table}

\begin{table}[h]
    \centering
    \tiny
    \resizebox{\linewidth}{!}{
    \begin{tabular}{c|c|c|c|c|c}
         \toprule
         model&backbone&resolution&params&speed&AP50\\
         \midrule
         Cascade&ResNet18&1333&56.11M&9.9&36.1 \\
         Cascade&ResNet18 & 2666&56.11M&5.4&40.3\\
         \hline
         Cascade&ResNet18&3800&56.11M&{\color{red}{2.9}}&{\color{blue}{42.6}}\\
         HRDNet&ResNet10+18&3800&62.44M&{\color{red}{3.7}}&49.2 \\
         HRDNet&ResNet18+101&3800&100.78M&2.8&53.3\\
         HRDNet&ResNeXt50+101&3800&152.22M&1.5&{\color{blue}{55.2}}\\
         \bottomrule
    \end{tabular}}
    \caption{The speed (items/second) and the number of parameters (M) are obtained on a same machine with one \textit{Nvidia GTX 2080Ti GPU} and \textit{Intel(R) Xeon(R) Silver 4210 CPU @ 2.20GHz}. The HRDNet here is a two stream version, using MS-FPN aligned with depth. }
    \label{tab:sizespeed}
\end{table}

\begin{table}[h]
    \centering
    \tiny
    \resizebox{\linewidth}{!}{
    \begin{tabular}{c|c|c|c|c|c}
         \toprule
         model&backbone&resolution&AP&AP50&AP75\\
         \midrule
         Single Backbone&ResNeXt50&3800&32.7&54.6&33.6 \\
         Single Backbone&ResNeXt101&1900&30.4&51.0&31.1 \\
         Model Ensemble&ResNeXt50+101&3800+1900&32.9&55.1&33.5 \\
         HRDNet&ResNeXt50+101&3800+1900&\textbf{33.5}&\textbf{56.3}&\textbf{34.0} \\ 
         
         \bottomrule
    \end{tabular}}
    \caption{\textbf{The comparison of HRDNet and Model Ensemble.} The models here follow the design of Cascade R-CNN. }
    \label{tab:multi_model_ensemble}
\end{table}
\vspace{-0.1cm}
\subsubsection{The effect of image resolution for small object detection}
Extensive ablation studies on the VisDrone2019 dataset are conducted to illustrate the effect of input image resolution for detection performance. Table \ref{tab:prove_size} shows that detection performance has a significant improvement with the increase of image resolution. Higher resolution leads to better performance under the same experimental settings. The performance of small objects presents more significant improvement from HRDNet. What is more, HRDNet performs much better than the state-of-the-art Cascade R-CNN with the same resolution as the input.


Interestingly, when the resolution of input increases, single backbone model, i.e. Cascade R-CNN, suffers dramatically decrease (~1.1-7.6\%) for categories with relatively large size, i.e. \textit{truck}, \textit{awning-tricycle} and \textit{bus}. On the contrary, significant performance increase (1-5.2\%) can be observed from HRDNet. Simply increasing the image resolution without considering the severe variant of object scale is not the ideal solution for object detection, let alone small object detection.

\vspace{-0.1cm}
\subsubsection{Explore the optimal image resolution}

We have stated and showed some experiments that the image resolution is important for small object detection; however, is it true higher resolution leads to better performance. Does it have the optimal resolution for detection? In this part, we will present the effect of image resolution for object detection. Figure \ref{fig:optim_size} shows the change of the Average Precise ($AP[0.05:0.95], AP50, AP75$) with different resolutions. The resolution starts from $2666$ (long edge) with 400 as the stride. Finally, HRDNet achieves the best performance when the resolution is $3800\times2800$ px. 

\vspace{-0.1cm}
\subsubsection{How to design the multi-scale FPN}
As mentioned above, MS-FPN is designed to fuse multi-scale feature groups. Here, we compared three different styles, including \textit{simple FPN, multi-scale FPN aligned by depth,  multi-scale FPN aligned by resolution}, as shown in Figure \ref{fig:compareFPN}, to demonstrate MS-FPN's advantage. A simple FPN is to apply standard FPN to each multi-scale group of HRDNet and finally fuse the results of each FPN. For multi-scale FPN, there are new connections between multi-streams, as shown in Figure \ref{fig:multiscalefpn}. We conducted two groups experiments with ResNet10+18 backbone and ResNet18+101 backbone. The first experiment in Table \ref{tab:multiscaleFPN} shows that the multi-scale FPN is better than the simple FPN. Both experiments demonstrate that MS-FPN aligned with depth performs better than those aligned with resolution. Therefore, we adopt MS-FPN aligned with depth in our architecture.

\begin{table*}[t]
    \tiny
    \centering
    \setlength{\tabcolsep}{5.6mm}{
    \begin{tabular}{c|c|c|c|c|c|c|c}
        \toprule
            \quad model&backbone&AP&AP50&AP75&$AP_{S}$&$AP_{M}$&$AP_{L}$\\
            \midrule
            R-FCN  \cite{dai2016rfcn}&ResNet-101&29.9&51.9&-&10.8&32.8&45.0\\
            Faster R-CNN w FPN  \cite{FPN}&ResNet-101&36.2&59.1&39.0&18.2&39.0&48.2\\
            DeNet-101(wide)  \cite{denet}&ResNet-101&33.8&53.4&36.1&12.3&36.1&50.8\\
            CoupleNet  \cite{couplenet}&ResNet-101&34.4&54.8&37.2&13.4&38.1&50.8\\
            Mask-RCNN  \cite{maskrcnn}&ResNeXt-101&39.8&62.3&43.4&22.1&43.2&51.2\\
            Cascade RCNN  \cite{cai2018cascade}&ResNet-101&42.8&62.1&46.3&23.7&45.5&55.2\\
            SNIP++  \cite{SNIP}&ResNet-101&43.4&65.5&48.4&27.2&46.5&54.9\\
            SNIPER(2scale)  \cite{SNIPER}&ResNet-101&43.3&63.7&48.6&27.1&44.7&56.1\\
            Grid-RCNN  \cite{gridrcnn}&ResNeXt-101&43.2&63.0&46.6&25.1&46.5&55.2\\
            
            \midrule
            SSD512  \cite{SSD}&VGG-16&28.8&48.5&30.3&10.9&31.8&43.5\\
            RetinaNet80++  \cite{retina}&ResNet-101&39.1&59.1&42.3&21.8&42.7&50.2\\
            RefineDet512  \cite{refinedet}&ResNet-101&36.4&57.5&39.5&16.6&39.9&51.4\\
            M2Det800&VGG-16&41.0&59.7&45.0&22.1&46.5&53.8\\
            \midrule
            CornetNet511  \cite{CornerNet}&Hourglass-104&40.5&56.5&43.1&19.4&42.7&53.9\\
            FCOS  \cite{tian2019fcos}&ResNeXt-101&42.1&62.1&45.2&25.6&44.9&52.0\\
            FSAF  \cite{fsaf}&ResNeXt-101&42.9&63.8&46.3&26.6&46.2&52.7\\
            CenterNet511  \cite{duan2019centernet}&Hourglass-104&44.9&62.4&48.1&25.6&47.4&57.4\\
            \midrule
            HRDNet++&ResNet101+152&\textbf{47.4}&\textbf{66.9}&\textbf{51.8}&\textbf{{\color{red}{32.1}}}&\textbf{50.5}&\textbf{55.8}  \\
        \bottomrule
    \end{tabular}}
    \caption{The state of the art of the performance on the MS COCO \textit{test-dev}, the input resolution of HRDNet ResNet101 stream is same as other models above, i.e. $1333\times800$, while the input of ResNet 152 stream is a $2\times$ smaller image. '++' denotes that the inference is performed with multi-scales etc.}
    \label{tab:cocosota}
\end{table*}

\begin{table*}[t]
    \tiny
    \centering
    \setlength{\tabcolsep}{4mm}
    \begin{tabular}{c|c|c|c|c|c|c|c|c|c}
    \toprule
    model&backbone&resolution&AP&AP50&AP75&AR1&AR10&AR100&AR500\\
    \midrule
    \dag Cascade R-CNN  \cite{cai2018cascade}&ResNet50&2666&24.10&42.90&23.60&0.40&2.30&21.00&35.20\\
    \dag Faster R-CNN  \cite{fasterrcnn}&ResNet50&2666&23.50&43.70&22.20&0.34&2.20&18.30&35.70\\
    \dag RetinaNet \cite{retina}&ResNet50&2666&15.10&27.70&14.30&0.17&1.30& 24.60&25.80\\
    \dag FCOS  \cite{tian2019fcos}&ResNet50&2666&16.60&28.80&16.70&0.38&2.20&24.40&24.40\\
    \hline
    
    HFEA  \cite{HFEA}&ResNeXt101&-&27.10&-&-\\
    HFEA  \cite{HFEA}&ResNeXt152&-&30.30&-&-\\
    DSOD  \cite{DSOD}&ResNet50&-&28.80&47.10&29.30\\
    \hline
    \dag HRDNet &ResNet10+18&3800&{\color{black}{28.68}}&49.15&28.90&0.48&3.42&37.47&37.47\\
    \dag HRDNet &ResNet18+101&{\color{black}{2666}}&{\color{black}{28.33}}&49.25&28.16&0.47&3.34&36.91&36.91\\
    \dag HRDNet &ResNet18+101&{\color{black}{3800}}&{\color{black}{31.39}}&53.29&31.63&0.49&3.55&40.45&40.45\\
    \dag HRDNet++                             & ResNet50+101+152 & 3800 & 34.35& 56.65 &35.51 &0.53 &4.00& 43.24 & 43.25 \\
    \dag HRDNet++                             & ResNeXt50+101 & 3800                  & \textbf{35.51}         & \textbf{62.00} & \textbf{35.13} & 0.39 & 3.38 & 30.91 &46.62 \\
    \bottomrule
    \end{tabular}
    \caption{\textbf{The comparison with the state-of-the-art object detection models on visdrone2019 DET validation set.} For DSOD results, we only show their true results without model ensemble. We only listed those results trained on VisDrone2019 train set. Those results with $\dag$ are trained and tested with the same environment and base code. '++' denotes that the inference is performed with multi-scales.}
    \label{tab:sota_visdrone}
\end{table*}
\begin{table}[t]
    \tiny
    \centering
    \setlength{\tabcolsep}{2mm}{
    \begin{tabular}{c|c|c|c}
        \toprule
            \quad model&backbone&input size&mAP\\
            \midrule
            Faster R-CNN  \cite{resnet}&ResNet-101&~$1000\times600$&76.4\\
            R-FCN  \cite{dai2016rfcn}&ResNet-101&~$1000\times600$&80.5\\
            OHEM  \cite{OHME}&VGG-16&~$1000\times600$&74.6\\
            HyperNet  \cite{hpernet}&VGG-16&~$1000\times600$&76.3\\
            R-FCN w DCN  \cite{deformable}&ResNet-101&~$1000\times600$&82.6\\
            CoupleNe  \cite{couplenet}t&ResNet-101&~$1000\times600$&82.7\\
            DeNet512(wide)  \cite{denet}&ResNet-101&~$512\times512$&77.1\\
            FPN-Reconfig  \cite{fpnreconfig}&ResNet-101&~$1000\times600$&82.4\\
            \midrule
            SSD512  \cite{SSD}&VGG-16&$512\times512$&79.8\\
            RefineDet512  \cite{refinedet}&VGG-16&$512\times512$&81.8\\
            RFBNet512  \cite{RFB}&VGG-16&$512\times512$&82.2\\
            CenterNet  \cite{objaspoint}&ResNet-101&$512\times512$&78.7\\
            CenterNet  \cite{objaspoint}&DLA  \cite{objaspoint}&$512\times512$&80.7\\
            \midrule
            HRDNet & ResNeXt50+101&$2000\times1200$&82.4\\
            HRDNet++ & ResNeXt50+101&$2000\times1200$&\textbf{83.4}\\
        \bottomrule
    \end{tabular}}
    \caption{The state of the art performance on Pascal VOC 2007 \textit{test}.}
    \label{tab:vocsota}
\end{table}
\vspace{-0.1cm}
\subsubsection{Efficient and Effective HRDNet}
HRDNet is a multi-streams network, and there may be some concerns about the model size and running speed. Here, we illustrate the number of parameters and running speed of our HRDNet, comparing with the state-of-the-art single backbone baseline. The results are shown in Table \ref{tab:sizespeed} demonstrate that our HRDNet can achieve much better performance with a similar number of parameters and even faster running speed. 



\vspace{-0.1cm}
\subsubsection{The comparison with single backbone model ensemble}
To further demonstrate that the performance improvement of HRDNet \textit{is not because of more parameters}, we compared two-stream HRDNet with the ensemble of two single backbone models under the same experimental setting (Table \ref{tab:multi_model_ensemble}). The ensemble models fuse the predicted bounding boxes and scores before NMS (Non-Maximum Suppression) and then perform NMS together. We found that the single backbone models with high-resolution input always perform better than those with low-resolution even it is processed by a stronger backbone. HRDNet performs better than the ensemble model, thanks to the novel multi-scale and multi-level fusion method. These results further prove that our designed MS-FPN is essential for HRDNet.

\vspace{-0.2cm}
\subsection{Comparison with the state-of-the-art methods}
\paragraph{VisDrone2019}
To demonstrate the advantage of our model and technical criterion, we also compare HRDNet with the most popular and state-of-the-art methods. Table \ref{tab:sota_visdrone} shows that our proposed HRDNet achieves the best performance on VisDrone2019 DET validation set. Notably, our model obtains more than 3.0\% AP improvement with ResNeXt50+101 compared to HFEA using ResNet152 as their backbone.

\paragraph{COCO2017}
Besides the experiments on VisDrone2019, we also conduct experiments on the COCO2017 test set to prove our method works well on a larger scale, complicated and standard detection dataset. Table \ref{tab:cocosota} shows that HRDNet achieves state-of-the-art results, and $>4.9\% AP_{small}$ improvement compared with most recent models.

\paragraph{Pascal VOC2007/2012}
There are not too many small objects in Pascal VOC. We conducted experiments on this data set to demonstrate HRDNet not only improves small object detection but also keeps the performance for large objects. 
\vspace{-0.3cm}
\section{Conclusion}
Merely increasing the image resolution without modifications will relatively damage the performance of large objects. Moreover, the server variance of object scales further limits the performance from high-resolution images. Motivated by this, we propose a new detection network for small objects, HRDNet. In order to handle the issues well, we further design MD-IPN and MS-FPN. HRDNet achieves the state-of-the-art on small object detection data set, VisDrone2019, at the same time, we outperform on other benchmarks, i.e., MS COCO, Pascal VOC. 



\bibliographystyle{named}
\bibliography{main.bib}

\end{document}